\documentclass[conference]{IEEEtran}
\IEEEoverridecommandlockouts
\usepackage{amsmath,amssymb,amsfonts}
\usepackage{algorithmic}
\usepackage{graphicx}
\usepackage{textcomp}
\usepackage{xcolor}
\def\BibTeX{{\rm B\kern-.05em{\sc i\kern-.025em b}\kern-.08em
    T\kern-.1667em\lower.7ex\hbox{E}\kern-.125emX}}
\usepackage{subcaption}
\usepackage[font=footnotesize, labelfont=bf]{caption}

\usepackage{makecell}
\usepackage{adjustbox}
\usepackage{longtable}
\usepackage{tabularx}
\usepackage[numbers]{natbib}
\usepackage{booktabs}
\usepackage{siunitx}
\usepackage{tabularx}
\usepackage{caption}
\usepackage{url}
\usepackage{threeparttable}
\title{WearableMil: An End-to-End Framework for Military Activity Recognition and Performance Monitoring}
\author{
\IEEEauthorblockN{Barak Gahtan}
\IEEEauthorblockA{%
  \textit{Department of Computer Science}\\
  \textit{Technion—Israel Institute of Technology}\\
  barakgahtan@cs.technion.ac.il}
\and
\IEEEauthorblockN{Shany Funk}
\IEEEauthorblockA{%
  \textit{Department of Information Systems}\\
  \textit{University of Haifa}\\
  shanyfit@gmail.com}
\and
\IEEEauthorblockN{Itay Ketko}
\IEEEauthorblockA{%
  \textit{Department of Information Systems}\\
  \textit{University of Haifa}\\
  iketko@gmail.com}
\and
\IEEEauthorblockN{Einat Kodesh}
\IEEEauthorblockA{%
  \textit{Department of Information Systems}\\
  \textit{University of Haifa}\\
  ekodesh@univ.haifa.ac.il}
\and
\IEEEauthorblockN{Tsvi Kuflik}
\IEEEauthorblockA{%
  \textit{Department of Information Systems}\\
  \textit{University of Haifa}\\
  tsvikak@univ.haifa.ac.il}
\and
\IEEEauthorblockN{Alex M.~Bronstein}
\IEEEauthorblockA{%
  \textit{Department of Computer Science}\\
  \textit{Technion—Israel Institute of Technology}\\
  bron@cs.technion.ac.il}
}

\begin{document}
\maketitle
\begin{abstract}
Musculoskeletal injuries during military training significantly impact readiness, making prevention through activity monitoring crucial. While Human Activity Recognition (HAR) using wearable devices offers promising solutions, it faces challenges in processing continuous data streams and recognizing diverse activities without predefined sessions. This paper introduces an end-to-end framework for preprocessing, analyzing, and recognizing activities from wearable data in military training contexts. Using data from 135 soldiers wearing \textit{Garmin--55} smartwatches over six months with over 15 million minutes. We develop a hierarchical deep learning approach that achieves 93.8\% accuracy in temporal splits and 83.8\% in cross-user evaluation. Our framework addresses missing data through physiologically-informed methods, reducing unknown sleep states from 40.38\% to 3.66\%. We demonstrate that while longer time windows (45-60 minutes) improve basic state classification, they present trade-offs in detecting fine-grained activities. Additionally, we introduce an intuitive visualization system that enables real-time comparison of individual performance against group metrics across multiple physiological indicators. This approach to activity recognition and performance monitoring provides military trainers with actionable insights for optimizing training programs and preventing injuries.
\end{abstract}
\begin{IEEEkeywords}
Human Activity Recognition, Deep Learning, Machine Learning, Physiologically Informed Imputation, Wearables Sensor Data, Injury Prevention, Group Visualization, Relative Ranking within Group
\end{IEEEkeywords}

\section{Introduction}
Military personnel often operate under extreme physical demands~\cite{10.1093/milmed/usac163} , making them highly susceptible to musculoskeletal injuries, which significantly impact operational readiness. Recent studies highlight the severity of this issue: up to 55\% of active component soldiers report injuries, with 72\% classified as overuse injuries~\cite{teyhen2018incidence}. These injuries result in substantial operational impacts through limited active-duty days~\cite{hauret2015epidemiology} and increased medical costs~\cite{teyhen2018incidence}.

Research has extensively explored risk factors and prevention strategies to mitigate these injuries, given their critical implications for military efficiency~\cite{molloy2020musculoskeletal}. Military organizations are increasingly adopting wearable technology for continuous monitoring and assessment~\cite{USDOD2023wearables}, with recent naval studies demonstrating the feasibility of large-scale physiological monitoring during operational exercises~\cite{wearableDevicesSailorSleep}. However, understanding soldiers' activities—critical to contextualizing and interpreting physiological data—remains a challenge, as activity information is not always clear or readily available.

Human Activity Recognition (HAR) has emerged as a key research area for addressing this challenge. By leveraging data from devices such as smartwatches equipped with accelerometers, gyroscopes, and heart rate (HR) monitors, HAR enables the automatic identification of activities, offering valuable insights into individual and group behaviors~\cite{zhang2022deep}. Military-specific implementations have shown promising results, achieving up to 80\% accuracy on controlled environments in real-time group activity recognition~\cite{mukherjee2017smartarm}. Recent advances in deep learning (DL) have further enhanced HAR by automating feature extraction and pattern recognition from raw sensor data~\cite{khatun2022deep}.

Despite these advancements, HAR systems face significant real-world challenges, particularly in group training or military scenarios. These include handling noisy and incomplete data, integrating heterogeneous sensor inputs, and ensuring robustness under varying environmental conditions~\cite{ngiam2011multimodal}. Moreover, capturing group dynamics and individual contributions within collective activities adds another layer of complexity. While existing HAR systems have shown promise in controlled environments, there remains a critical need for robust frameworks that can handle the complexity and scale of military training scenarios while providing actionable insights for injury prevention and performance optimization~\cite{nindl2015human}.

This paper introduces an end-to-end framework for preprocessing, analyzing, and recognizing activities from large-scale military training datasets. Our approach addresses missing data through physiological insights and unifies multimodal data via a novel Linear Truncated Model (LTM). We validate our framework using data from 135 soldiers wearing \textit{Garmin 55} smartwatches over six months (15 million data-minutes), encompassing HR, physical activities, sleep data, and daily schedules. Our experiments demonstrate the framework's effectiveness in identifying activities such as running, obstacle training, and firearm training across both temporal and out-of-sample evaluations. Additionally, we present a visualization tool highlighting individual deviations from group norms through key metrics (pulse rate, pulse-to-min/max ratios, steps and distance per minute).

Our key contributions include: (1) a novel physiologically-informed imputation strategy that leverages domain knowledge to handle missing data in multimodal sensor streams, (2) a hierarchical DL architecture optimized for military activity recognition that handles both coarse and fine-grained activity classifications, and (3) a scalable approach to group-context analysis that enables real-time performance monitoring across military units.

The rest of the paper is organized as follows: Section~\ref{section2} reviews related work, Section~\ref{section3} details preprocessing and imputation, and Section~\ref{section344} presents the model design and loss function. Section~\ref{section4} evaluates the framework's performance under two scenarios: temporal splits and cross-user evaluation. Section~\ref{section5} introduces our relative visualization tool for group performance analysis, and Section~\ref{section6} discusses the framework's limitations and concludes with future research directions. \label{section1}

\section{Related Work}
The prevention of musculoskeletal injuries among military personnel has long been a focal point of research due to its significant impact on operational readiness~\cite{khatun2022deep}. Recent studies highlight that up to 55\% of active component soldiers report injuries, with 72\% classified as overuse injuries~\cite{teyhen2018incidence}. Understanding the training context is a critical first step in analyzing soldier behavior and performance to identify and mitigate injury risks. HAR provides a powerful framework for addressing this challenge, with military organizations increasingly adopting wearable technology for continuous monitoring and assessment~\cite{USDOD2023wearables}.

To address these challenges, HAR systems have gained prominence across diverse domains, including healthcare, rehabilitation, and surveillance systems, by leveraging wearable sensors such as accelerometers, gyroscopes, and magnetometers~\cite{challa2022multibranch}. In military contexts specifically, implementations like SmartARM have achieved 80\% accuracy in real-time group activity recognition~\cite{mukherjee2017smartarm}, demonstrating the practical viability of these approaches in operational environments.

Building on these foundations, recent advances in DL have transformed HAR by automating feature extraction and improving recognition accuracy. Techniques like hybrid CNN-LSTM models integrated with self-attention and multibranch CNN-BiLSTM architectures excel in capturing both local and long-term dependencies in sequential data~\cite{khatun2022deep}. However, despite these advancements, real-world applications of HAR continue to face challenges such as device placement variability, environmental noise, and imbalanced datasets. Strategies like ensemble learning and data augmentation have shown promise in enhancing model robustness~\cite{zhao2021data}.

These challenges become particularly evident in group-level HAR, an emerging subfield that introduces additional complexities such as recognizing collective actions and evaluating individual performance within group contexts. Recent naval studies have demonstrated the feasibility of large-scale physiological monitoring, with over 200 sailors successfully monitored during operational exercises~\cite{wearableDevicesSailorSleep}. While techniques combining neural networks and IMU sensors have demonstrated high accuracy in group activity recognition~\cite{mao2023hybrid}, graphical models, such as active factor graph networks, further enhance this capability by capturing interactions between group members and providing insights into collective dynamics~\cite{xie2024active}.

A fundamental challenge across these applications, especially in multimodal settings, is addressing both noise and missing data. Studies comparing consumer wearables to clinical sleep assessment tools have shown promising results, with sensitivity between 0.883 and 0.977 for sleep-wake classification~\cite{roberts2020detecting}. While simple imputation methods (e.g., zero-filling, linear interpolation) can disrupt cross-sensor correlations, more complex models like UniTS~\cite{ngiam2011multimodal} attempt to handle these issues but often suffer from poor generalization and high computational complexity. This underscores the need for preprocessing pipelines that incorporate domain knowledge to address these challenges effectively.

The application of wearable sensors extends beyond basic HAR to play a vital role in intelligent healthcare systems, where multisensory fusion has enhanced diagnostic and monitoring capabilities~\cite{ma2018intelligent}. A prime example is the U.S. Army's Heat Injury Prevention System (HIPS), which demonstrates the potential of real-time monitoring for injury prevention by successfully predicting heat-related injuries through combined physiological metrics and movement patterns~\cite{georgiaTechHeatIllness,gtriWearableSensorSystem}. This integration of multiple data sources is further supported by studies leveraging HR variability~\cite{luo2021spectral}, sleep patterns~\cite{kleinsasser2022detecting}, and training logs~\cite{lovdal2021injury}, which highlight the importance of combining physiological and activity data for comprehensive injury risk assessment.

Building upon these advances and addressing the identified challenges, this study introduces an end-to-end preprocessing pipeline for handling noise and missing data across multimodal wearable datasets. By leveraging physiological patterns for imputation and unifying data onto a common grid, our approach enables accurate activity recognition and group-level performance evaluation without the need for explicitly labeled training sessions. 
\label{section2}

\section{Data Processing and Preparation Framework}
\begin{figure}[t]
    \centering
    \includegraphics[width=1\linewidth]{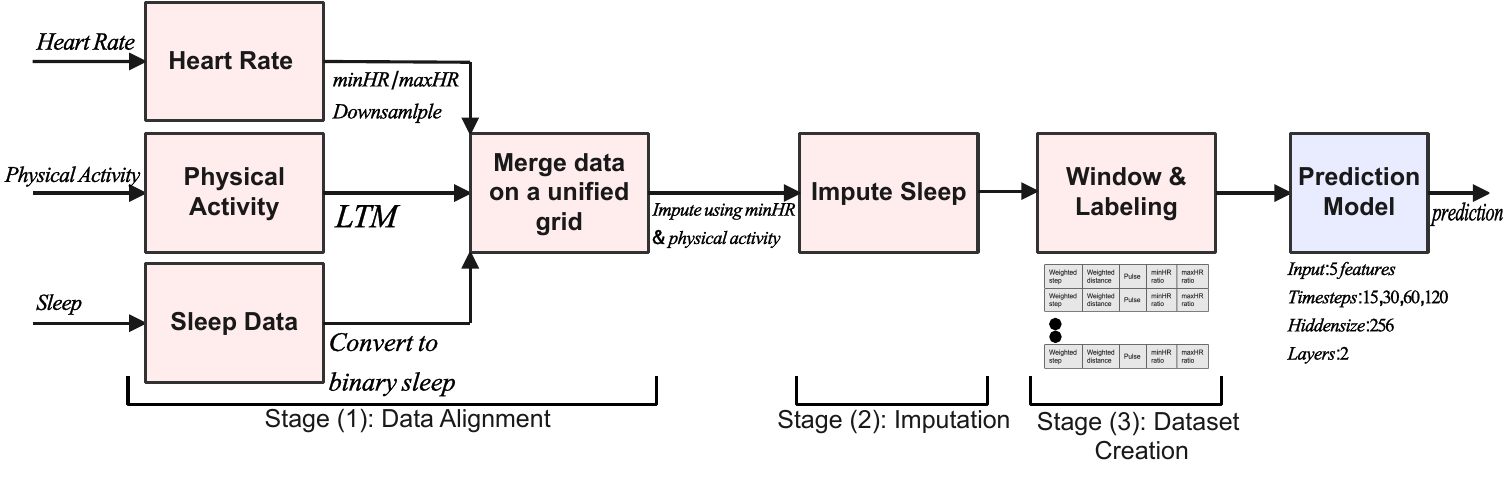}
    \caption{Overview of the Data Processing and Prediction Pipeline. The pipeline integrates multi-source wearable data, including HR, physical activity, and sleep data, into a unified grid using the LTM. Missing sleep data is imputed leveraging physiological insights, HR, and physical activity thresholds. Windows of varying sizes are labeled and inputted into a neural network for sequential activity prediction.}
    \label{fig:pipeline}
\end{figure}

Figure~\ref{fig:pipeline} outlines the data processing pipeline used to transform wearable device data into a format suitable for supervised learning. The dataset was collected from 135 soldiers using \textit{Garmin Forerunner 55} devices, spanning six months of continuous recording with over 15 million data-
minutes points~\citep{ortiz2024data}. \footnote{Ethical approval was granted by the Helsinki Committee of the IDF’s Medical Corps (1998-2019) and the Human Use Committee of Sheba Medical Center (6333-19-SMC). All participants signed informed consent.}
The pipeline comprises three main stages: data alignment on a unified time grid, imputation of missing sleep states, and dataset creation for ML.

\textbf{Data Alignment.} 
HR, physical activity (steps and distance), and sleep data streams were aligned onto a unified one-minute grid~\citep{mandadapu2023automated}. HR data, originally sampled at 15-second intervals, was downsampled by averaging. Personalized HR thresholds (\textit{minHR} and \textit{maxHR}) were computed using the 5th and 99.97th percentiles of daily HR values based on physiological literature~\citep{baust1969regulation}.

The LTM redistributed activity data from 15-minute intervals to our one-minute grid, operating on the assumption that physical activity intensity correlates with HR variations within each time block~\citep{schrack2018using}. Steps and distance were allocated proportionally based on the HR of each minute within the 15-minute block. An HR cutoff of 1.05 $\times$ \textit{minHR} was applied to exclude noise and low-intensity activity, stricter than the 20-30\% range typically used in literature~\citep{schlagintweit2023effects, chandola2013sleep}.

\textbf{Imputation.} Missing sleep states were imputed using a rule-based framework guided by HR and physical activity thresholds:
\begin{enumerate}
    \item Rule 1 (Sleep Classification): Minutes with HR below $1.05 \times \textit{minHR}$ and zero steps during nighttime were classified as ``Sleep''. For daytime naps, the threshold was adjusted to $1.2 \times \textit{minHR}$.
    \item Rule 2 (Awake Classification): Minutes with HR exceeding $1.2 \times \textit{minHR}$ or non-zero steps were classified as awake.
    \item Rule 3 (Transitions): Missing states flanked by identical known states were classified based on surrounding context.
\end{enumerate}

\begin{table}[t]
\scriptsize	
\centering
\caption{Impact of HR-Based Sleep-Wake Imputation Rules. Three-stage rule-based framework using HR thresholds and activity data.}
\label{tab:imputation_impact}
\setlength{\tabcolsep}{3pt}
\begin{tabular}{lrrrr}
\toprule
Metric & Pre-Imput. & After 1+2$^1$ & After 1+2+3$^2$ & Net \\
\midrule
\multicolumn{5}{l}{\textit{Per-User Avg. (Min)}} \\ 
Total Min & 109,618 & 109,618 & 109,618 & -- \\
Sleep Min & 63,011 & 65,716 & 65,723 & +2,712 \\
Awake Min & 2,342 & 50,231 & 50,236 & +47,894 \\
Unknown Min & 44,265 & 4,013 & 4,008 & -40,257 \\
\midrule
\multicolumn{5}{l}{\textit{Classification (\%)}} \\
Sleep$^a$ & 57.48 & 59.95 & 59.96 & +2.48 \\
Awake$^b$ & 2.14 & 45.82 & 45.83 & +43.69 \\
Unknown$^c$ & 40.38 & 3.66 & 3.66 & -36.72 \\
\midrule
\multicolumn{5}{l}{\textit{Rule Application (Min)}} \\
Rule 1 HR+Sleep & -- & 2,341 & 2,341 & 2,341 \\
Rule 2 HR+Awake & -- & 44,265 & 44,265 & 44,265 \\
Rule 3 Transit. & -- & -- & 24 & 24 \\
\bottomrule
\multicolumn{5}{l}{\tiny $^a$Sleep=(Sleep/Total)×100; $^b$Awake=(Awake/Total)×100;}\\[-2pt]
\multicolumn{5}{l}{\tiny $^c$Unknown=(Unknown/Total)×100; $^1$Rules 1+2=Sleep/Awake;}\\[-2pt]
\multicolumn{5}{l}{\tiny $^2$Rule 3=Transitions}
\end{tabular}
\end{table}

As shown in Table~\ref{tab:imputation_impact}, this approach significantly reduced unknown states from 40.38\% to 3.66\% of total time. Rule 1 classified sleep periods by identifying minutes with HR below $1.05 \times \textit{minHR}$ during nighttime (adjusted to $1.2 \times \textit{minHR}$ for daytime naps), contributing 2,341 minutes per user. Rule 2 had the largest impact, classifying 44,265 minutes per user as awake when HR exceeded $1.2 \times \textit{minHR}$ or steps were recorded. Finally, Rule 3 resolved transitions between known states. The framework increased identified sleep time by 2.48\%, representing an additional 2,712 minutes per soldier. This improvement is particularly significant in the military context, where soldiers often need to sleep during non-traditional hours due to operational demands. The ability to detect these irregular sleep periods is crucial for monitoring soldier fatigue and readiness. Overall, the framework substantially improved wake detection from 2.14\% to 45.83\% of total time while maintaining physiologically plausible sleep patterns.

\textbf{Unified Dataset Creation.} After alignment and imputation, HR features (\textit{pulse, pulse-to-minHR, pulse-to-maxHR}), steps, distance, and sleep states were merged into a unified dataset with sliding windows of \textbf{15, 30, 45, and 60 minutes} with \textbf{30\% overlap}. These window lengths were selected based on typical military activity durations and the need to capture both short-term variations and sustained patterns. Each window was assigned an activity label only if that activity occupied at least \textbf{70\%} of the window duration.

The labeling scheme follows a two-level hierarchy~\citep{fazli2021hhar}: Level 1 includes \textbf{Sleep}, \textbf{Awake}, and \textbf{Activity} (Other), while Level 2 refines \textbf{Activity} into specific subcategories including \textbf{Firearms Training}, \textbf{Military Drills}, \textbf{Running Exercise}, \textbf{Obstacle Course Training}, and others. To handle computational load, we applied stratified sampling with rates of \textbf{15\%} for 15-minute windows, \textbf{25\%} for 30-minute and 45-minute windows, and \textbf{40\%} for 60-minute windows. Class imbalance was addressed by generating synthetic samples for minority classes with Gaussian noise (SD: \textbf{0.03\%}).\label{section3}

\section{Model Design and Loss Function}
\textbf{Model Architecture}
Our two-tiered model architecture integrates a bidirectional LSTM network with specialized classifiers for hierarchical prediction~\cite{murtagh2012algorithms}. The model consists of: (1) a two-layer bidirectional LSTM (hidden dim=256, dropout=0.1) for temporal encoding, and (2) two specialized classifiers for Level 1 (\textbf{Sleep}, \textbf{Awake}, \textbf{Activity}) and Level 2 (specific activities) classification.

\textbf{Hierarchical Focal Loss}
We developed a custom loss function combining level-specific losses with configurable weights:: $
\mathcal{L}_{\text{total}} = \lambda_1 \mathcal{L}_{\text{level1}} + \lambda_2 \mathcal{L}_{\text{level2}}
$. Through an ablation study comparing combinations of \(\lambda_1 = \{0.1,0.3,0.5\}\) and \(\lambda_2 = \{1.0, 1.3, 1.5\}\), we determined optimal weights of \(\lambda_1 = 0.3\) and \(\lambda_2 = 1.0\). For each hierarchical level, we employ a focal loss formulation~\citep{yeung2022unified,ross2017focal}: $\mathcal{L}_{\text{focal}} = -\alpha (1 - p_t)^\gamma \log(p_t)
$
where \(p_t = \exp(-\mathcal{L}_{\text{CE}})\) with \(\mathcal{L}_{\text{CE}}\) being the cross-entropy loss. We set \(\alpha = 2.0\) to scale the loss for class imbalance and \(\gamma = 2.0\) to focus training on hard-to-classify samples. The level-specific losses are computed as:
$
\mathcal{L}_{\mathrm{level_{1,2}}} = \mathbb{E}[\mathcal{L}_{\mathrm{focal}}(\hat{y}_{\mathrm{level_{1,2}}}, y_{\mathrm{level_{1,2}}})]
$, where \(\hat{y}\) and \(y\) represent predicted and true labels respectively.\label{section344}

\section{Experimental Evaluation}
We evaluated our model under two scenarios~\citep{stojchevska2023lab}: (1) \textbf{Temporal Split} (70/15/15\% of each soldier's days for train/val/test) to assess within-user generalization, and (2) \textbf{User Split} (70/15/15\% of soldiers) to evaluate cross-user generalization. Tests spanned multiple window sizes (15--60 minutes) and both classification levels, using accuracy, F1-score, and ROC AUC metrics. Training employed AdamW optimizer~\citep{loshchilov2017fixing} (batch size=256, lr=1e-3, weight decay=0.01) with ReduceLROnPlateau scheduling and early stopping. The model was trained using a batch size of 256 with a learning rate scheduler \texttt{ReduceLROnPlateau}~\citep{smith2017cyclical}, and the AdamW optimizer~\citep{loshchilov2017fixing} with a learning rate of 1e-3 and weight decay of $0.01$ to mitigate overfitting.  
\begin{figure}[t]
\centering
\includegraphics[width=0.95\linewidth]{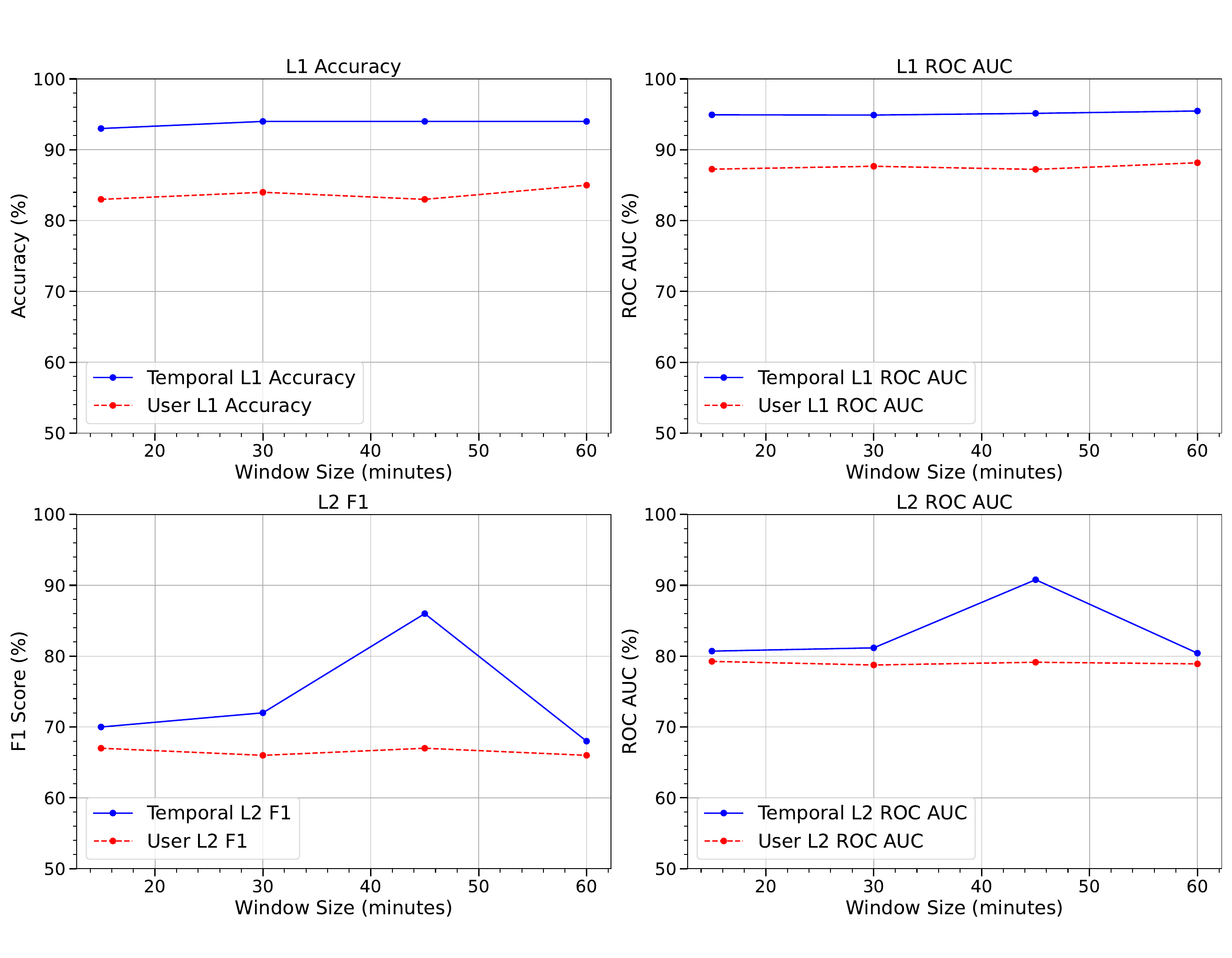}
\caption{\textbf{Performance Trends Across Window Sizes.} The 2x2 grid compares Level 1 (L1) and Level 2 (L2) metrics for temporal (blue) and user (red) splits across varying window sizes. Temporal splits consistently outperform user splits across all metrics. L1 metrics (top row) remain stable for temporal splits, while user splits exhibit variability. For L2 metrics (bottom row), temporal splits peak at 45 minutes, particularly in F1 score and ROC AUC, before declining slightly at 60 minutes. User splits show stable but lower performance due to challenges in generalizing to unseen users. The uniform y-axis range (50--100\%) ensures comparability across subplots.}
\label{fig:performance_trends}
\end{figure}   

\textbf{Temporal vs. User Generalization} Temporal splits consistently outperform user splits across all metrics, as shown in Table~\ref{tab:split_comparison}. This performance gap is particularly evident in Level 1 accuracy (93.8\% vs 83.8\%) and Level 2 F1-score (74.0\% vs 66.5\%). Figure~\ref{fig:performance_trends} further illustrates the model's performance across window sizes and evaluation scenarios. This performance gap widens with increasing window sizes, particularly for Level 2 metrics, highlighting the model's stronger ability to learn within-user patterns compared to generalizing across different users. The analysis reveals several key patterns: Level 1 metrics remain relatively stable across temporal splits as window size varies, whereas Level 2 metrics show a distinctive peak at 45 minutes, suggesting this \textbf{window length} provides an optimal balance between contextual information and sufficient class samples. \textbf{Class Distribution Effects:} The decline in Level 2 performance at larger window sizes arises from the 70\% majority label criterion, significantly reducing sample counts for certain activities; for example, ``Military Drills'' drops from 3,705 samples at 15 minutes to just 570 at 60 minutes, and activities with short durations like ``Wake Up'' (300 samples at 15 minutes) disappear entirely at larger windows due to their brief scheduled duration of 20 minutes. \textbf{Cross-User} challenges further exacerbate this trend, exemplified by activities such as ``Running Exercise'', which decreases sharply from 172 samples to 31 samples in the user split scenario as window sizes increase. Overall, temporal splits effectively capture within-user patterns, whereas user splits consistently struggle due to inter-user variability and data imbalance.

\begin{table}[t]
\footnotesize
\centering
\caption{Performance Comparison between Temporal and User Splits across Classification Levels}
\label{tab:split_comparison}
\begin{tabularx}{\columnwidth}{lXXXXX}
\toprule
\textbf{Split Type} & \textbf{Accuracy  (L1)} & \textbf{F1 (L1)} & \textbf{AUC (L1)} & \textbf{F1 (L2)} & \textbf{AUC (L2)} \\ \midrule
Temporal           & 93.8\%               & 93.8\%          & 95.3\%               & 74.0\%          & 84.1\%               \\
User               & 83.8\%               & 83.8\%          & 87.6\%               & 66.5\%          & 79.0\%               \\ \bottomrule
\end{tabularx}
\end{table} 
Overall, the performance trends reveal the following: (1) Larger windows generally improve Level 1 metrics due to richer contextual information; (2) Level 2 metrics are more sensitive to class distribution and the stricter majority label threshold imposed by larger windows, leading to a decline at 60 minutes; (3) Temporal splits benefit from temporal continuity, achieving significantly higher metrics across all levels compared to user splits, which face challenges from unseen user variability.
\begin{figure*}[t]
\centering
\includegraphics[width=0.70\linewidth]{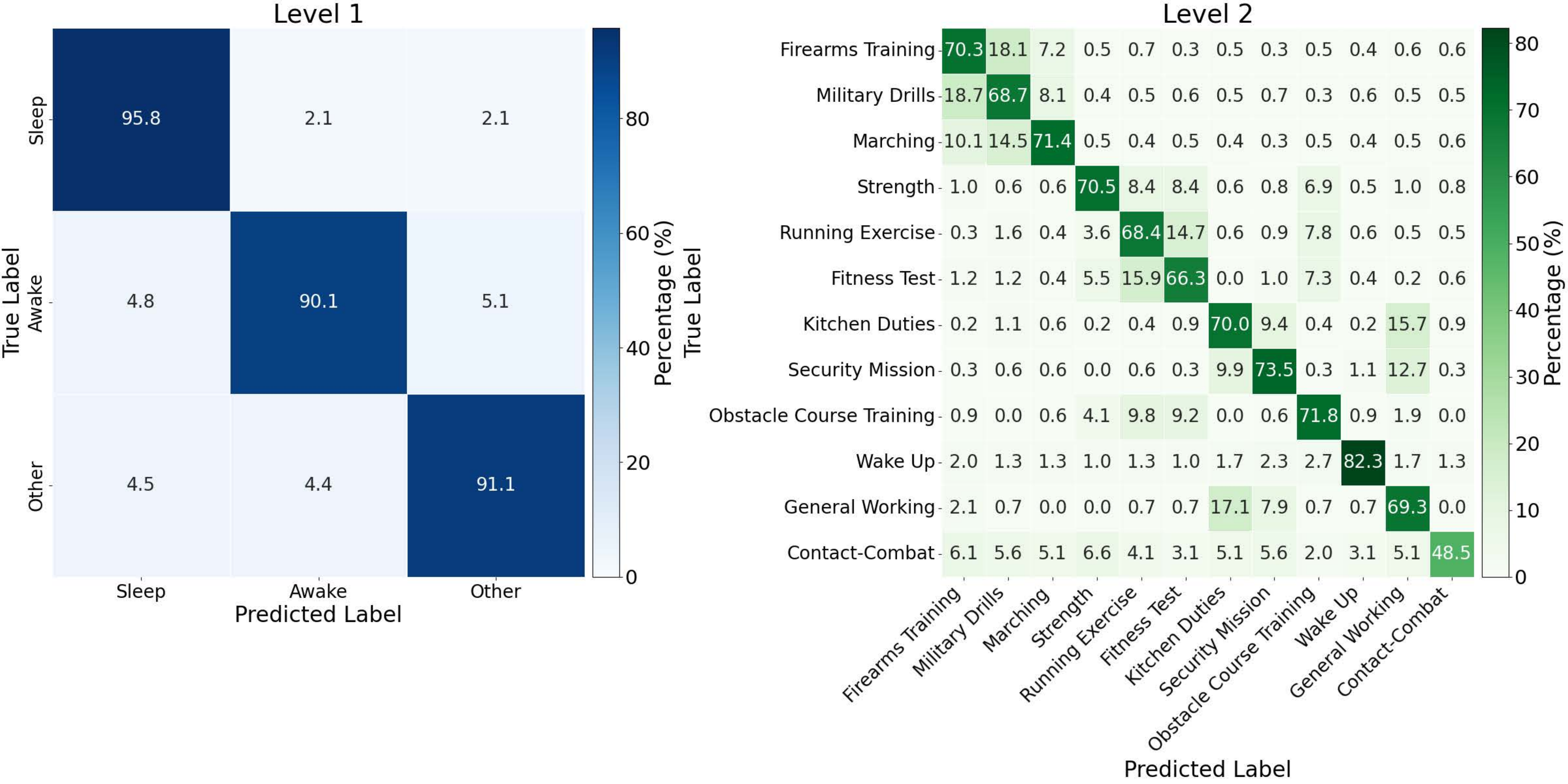}
\caption{Confusion matrix for the 30-minute temporal split (Level 1 and 2).}
\label{fig:confusion_matrix}
\end{figure*} 

Figure~\ref{fig:confusion_matrix} presents confusion matrices for Level 1 and Level 2 classification under the 30-minute temporal split. Level 1 classification performs strongly across all activities, notably ``Sleep'' (95.8\%). Level 2 analysis reveals several patterns: activities with distinct temporal or physiological signatures like ``Wake Up'' (82.3\%) and ``Security Mission'' (73.5\%) achieve high accuracy. However, physically similar activities show notable confusion, such as ``Military Drills'' misclassified as ``Firearms Training'' (18.7\%) and ``Running Exercise'' confused with ``Fitness Test'' (14.7\%), due to similar intensity and movement patterns. Activities like ``Contact-Combat'' exhibit lower accuracy (48.5\%), potentially because soldiers remove watches, while ``General Working'' and ``Kitchen Duties'' frequently overlap (15.7\% mutual confusion) due to similar low-intensity physical activity patterns.

The confusion matrix highlights the importance of distinct feature sets and balanced class representation to reduce misclassification driven by physical similarities and contextual overlaps. Overall, the framework demonstrates strong performance in recognizing both coarse and fine-grained military activities, particularly within individuals' patterns \textbf{over time}. High accuracy in sleep detection ensures reliable rest monitoring, while differentiation between training activities enables detailed activity tracking despite physical similarities. Superior performance in temporal splits supports the model’s effectiveness in longitudinal monitoring, although generalization to new soldiers remains challenging. These capabilities facilitate early injury intervention by continuously tracking physical strain from high-intensity activities (e.g., ``Military Drills'' at 68.7\%, ``Running Exercise'' at 68.4\%) and accurately detecting rest periods. Collectively, these features support operational readiness through improved training management and fatigue prevention.\label{section4}

\section{``Smart'' visualization for an individual}
Developing smart visualization tools is essential for analyzing individual physical performance relative to group norms in military training contexts~\citep{10.1093/milmed/usac163,murdock2018soldier}. These tools facilitate intuitive performance assessment relative to group norms, fair benchmarking, personalized training planning, and early detection of performance issues, crucial for injury prevention and operational readiness.

\begin{figure*}[t]
    \centering
    \begin{subfigure}{0.32\textwidth}
        \centering
        \includegraphics[width=\textwidth]{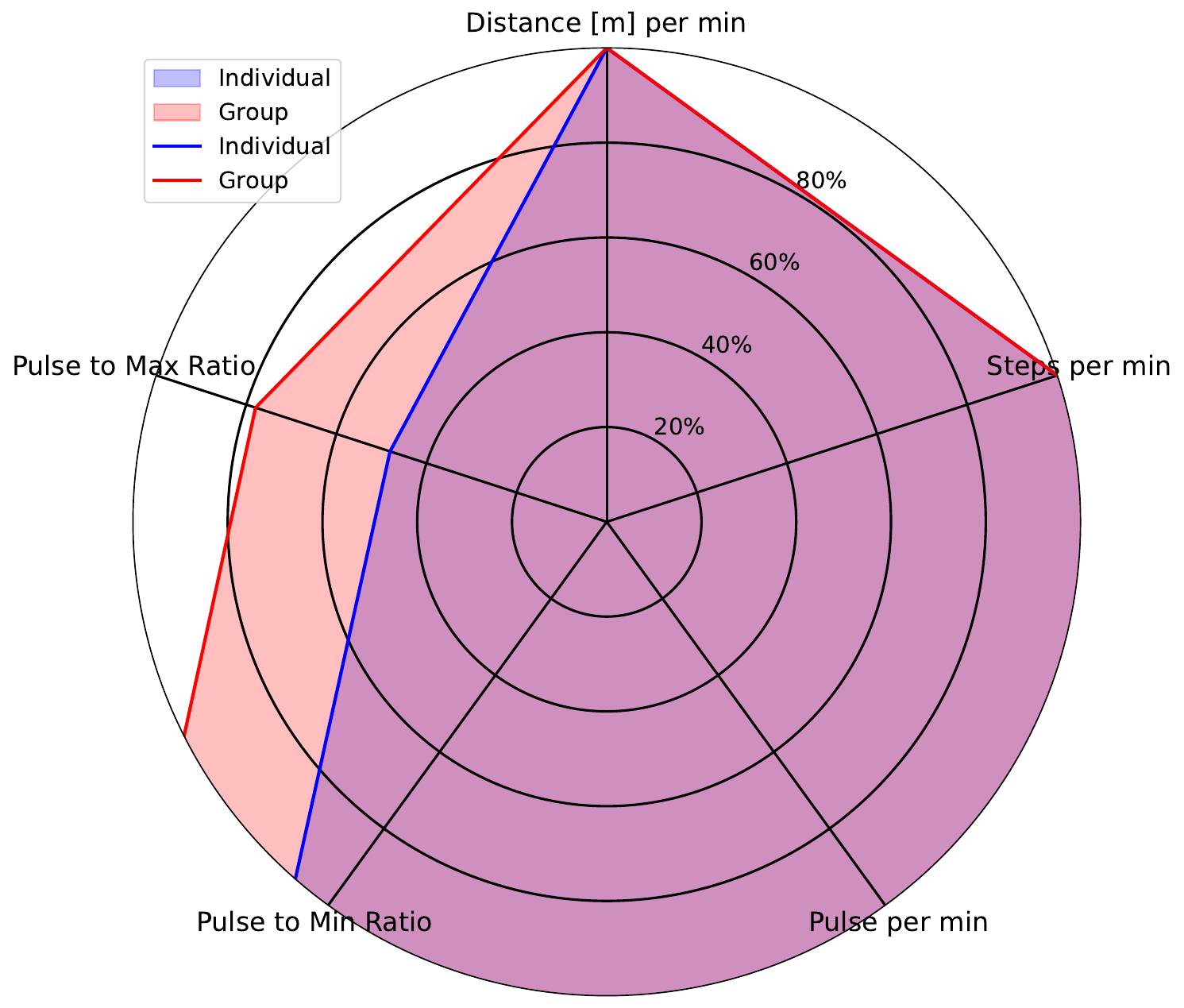}
        \caption{``Fitness Test''}
        \label{fig:fitness-test}
    \end{subfigure}%
    \hspace{1mm}
    \begin{subfigure}{0.32\textwidth}
        \centering
        \includegraphics[width=\textwidth]{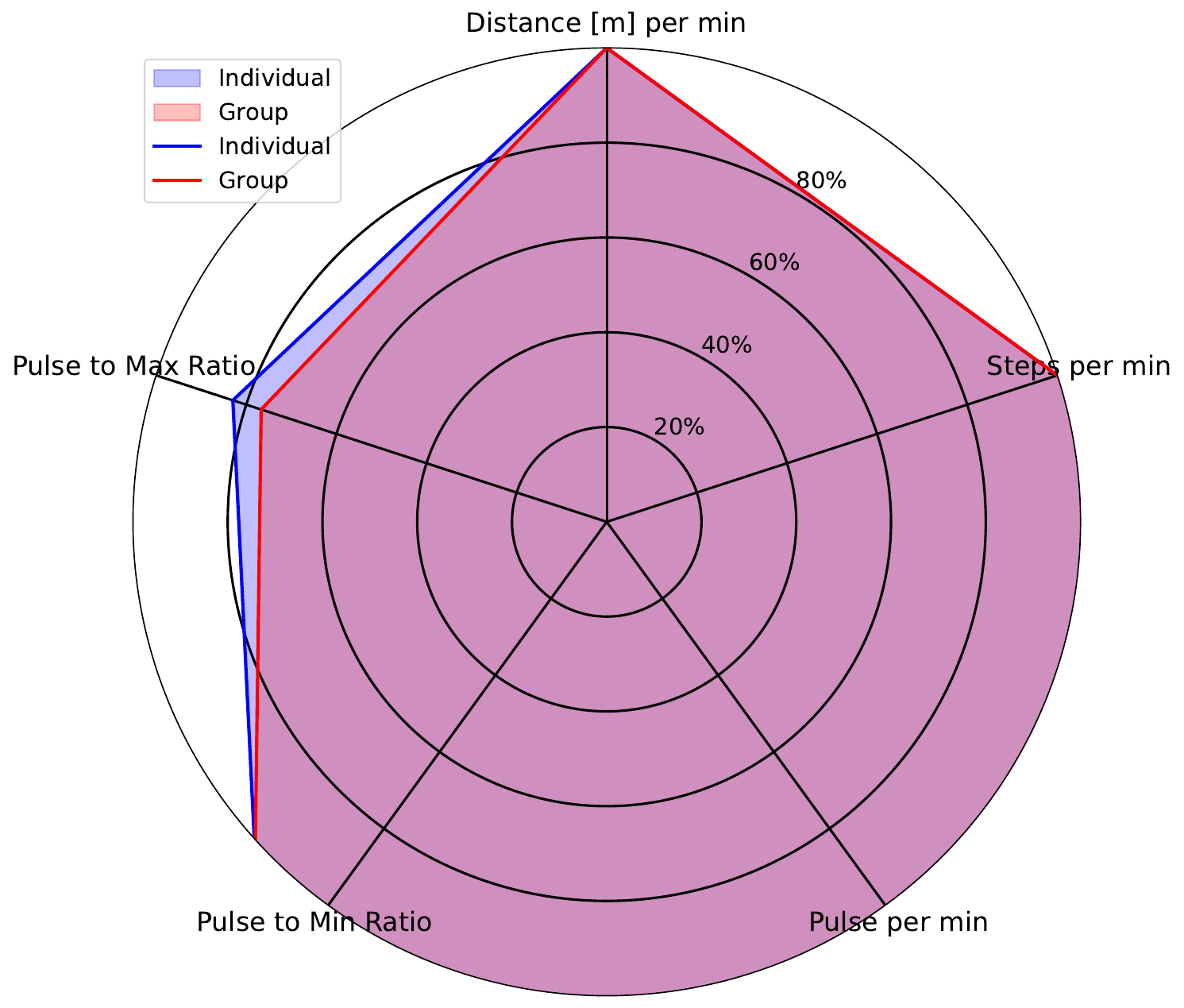}
        \caption{``Military Drill''}
        \label{fig:military-drill}
    \end{subfigure}
    \hspace{1mm}
    \begin{subfigure}{0.32\textwidth}
        \centering
        \includegraphics[width=\textwidth]{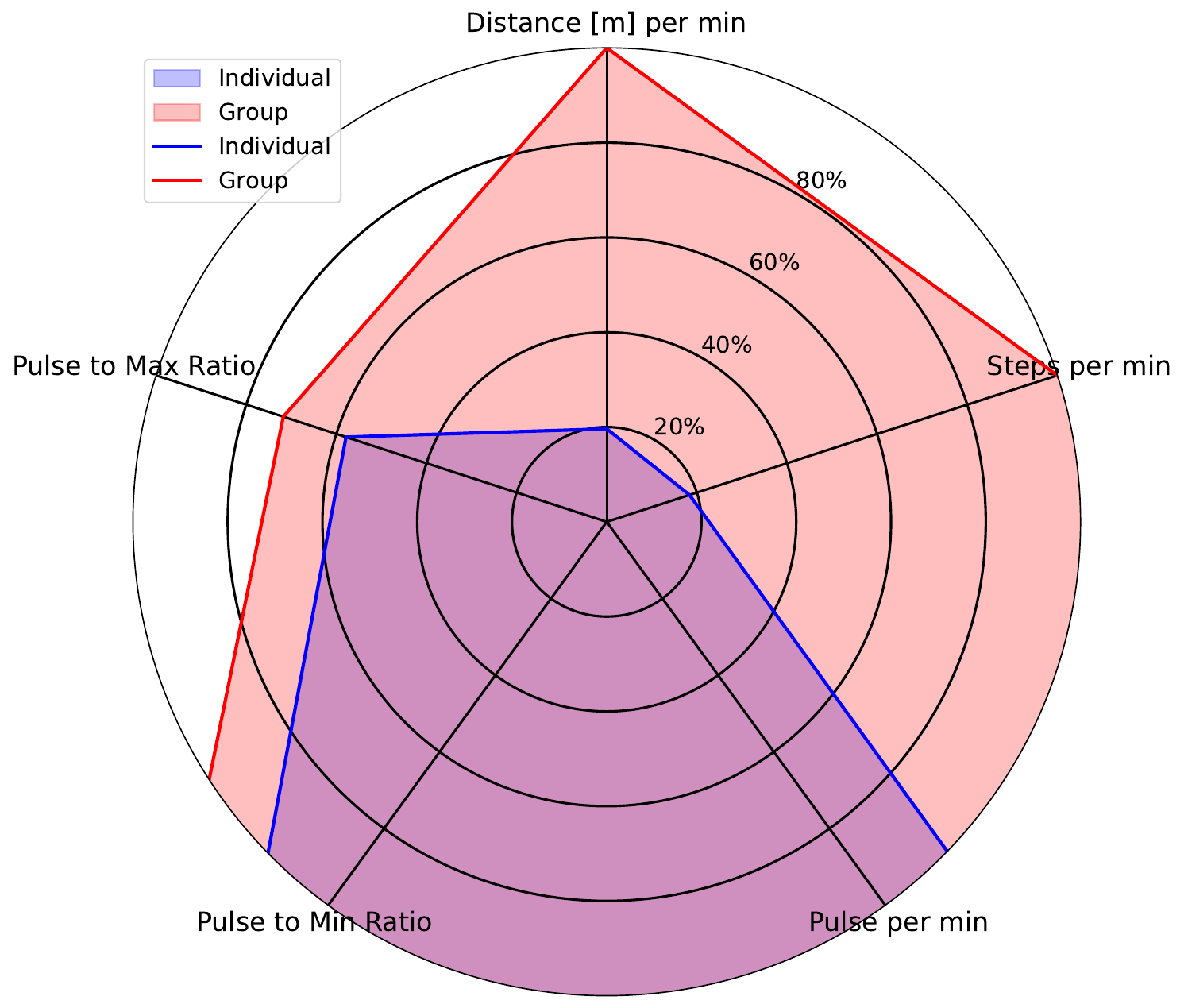}
        \caption{``Firearms Training''}
        \label{fig:rifling}
    \end{subfigure}%
    \caption{Individual performance across tasks (a-c), represented by the blue line, is compared to the group median (red line) across multiple metrics, including distance per minute, steps per minute, and pulse ratios. Solid lines indicate median values, while shaded areas show variability (data range) for each metric, reflecting consistency—larger shaded areas indicate higher variability.}
    \label{fig:performance_comparison}
\end{figure*}

Figure~\ref{fig:performance_comparison} shows radar charts comparing individual performance against group medians in three activities: ``Fitness Test'', ``Military Drill'', and ``Firearms Training''. Five key metrics (\textit{distance per minute}, \textit{steps per minute}, \textit{pulse per minute}, \textit{pulse to min ratio}, and \textit{pulse to max ratio}) are normalized (0–100\%) for consistency. Individual and group median performances are depicted by solid blue and red lines, respectively, with shaded areas representing variability (one standard deviation).

Performance metrics are interpreted through ratios that provide insight into effort and efficiency~\citep{mongin2020complex}. A higher \textit{pulse to max ratio} indicates greater cardiovascular exertion. Conversely, a lower \textit{pulse to min ratio} suggests better efficiency, with current pulse closer to resting levels. These ratios help distinguish between effort intensity and cardiovascular efficiency during activities.

The radar charts highlight distinct activity-specific performance patterns. In the ``Fitness Test'' (Figure~\ref{fig:performance_comparison}(a)), the individual matches group distance and steps but demonstrates better cardiovascular efficiency (lower pulse-to-max ratio). Conversely, during the ``Military Drill'' (Figure~\ref{fig:performance_comparison}(b)), the individual experiences higher cardiovascular strain despite similar physical output, suggesting potential conditioning needs. The individual notably underperforms in ``Firearms Training'' (Figure~\ref{fig:performance_comparison}(c)), with reduced physical activity and cardiovascular engagement, requiring immediate intervention.

These visualizations support proactive training management by enabling supervisors to rapidly detect anomalies (identify individuals who deviate significantly from group norms), adjust training intensities, develop targeted interventions, and effectively monitor training loads to minimize injury risks and maximize performance outcomes.\label{section5}


\section{Discussion, Conclusions, and Future Work}
Our framework reveals key trade-offs in wearable-based military activity recognition while achieving strong performance (93.8\% accuracy for Level 1, 74.0\% F1-score for Level 2) in both temporal and cross-user scenarios. Longer windows (45-60 min) improve basic state classification but may miss brief critical events like ``Wake Up'' (20 min) or short ``Running Exercise'' sessions essential for assessing training intensity. Our physiologically-informed sleep imputation significantly reduces unknown states (from 40.38\% to 3.66\%), enabling comprehensive recovery analysis crucial for soldier readiness.

The framework enhances military training through: (1) real-time visualization via radar charts comparing multiple metrics simultaneously (pulse rate, pulse-to-min/max ratios, and LTM-derived measures), providing supervisors actionable insights for detecting performance anomalies, and (2) individualized training optimization and injury prevention. Limitations include reliance on statistical HR calculations rather than ML methods, fixed window sizes potentially missing short-duration activities, and technical challenges from sensor inaccuracies and device removal during specific activities (e.g., ``Contact-Combat,'' 48.5\% accuracy).

Future work includes exploring advanced ML approaches like generative adversarial networks for improved data imputation, implementing adaptive window sizes to optimize fine-grained recognition, and integrating injury prediction models leveraging activity and sleep patterns. \label{section6}

{\scriptsize	
\bibliographystyle{IEEEtran}{}
\bibliography{refs}}

\begin{thebibliography}{10}
\providecommand{\url}[1]{#1}
\csname url@samestyle\endcsname
\providecommand{\newblock}{\relax}
\providecommand{\bibinfo}[2]{#2}
\providecommand{\BIBentrySTDinterwordspacing}{\spaceskip=0pt\relax}
\providecommand{\BIBentryALTinterwordstretchfactor}{4}
\providecommand{\BIBentryALTinterwordspacing}{\spaceskip=\fontdimen2\font plus
\BIBentryALTinterwordstretchfactor\fontdimen3\font minus \fontdimen4\font\relax}
\providecommand{\BIBforeignlanguage}[2]{{%
\expandafter\ifx\csname l@#1\endcsname\relax
\typeout{** WARNING: IEEEtran.bst: No hyphenation pattern has been}%
\typeout{** loaded for the language `#1'. Using the pattern for}%
\typeout{** the default language instead.}%
\else
\language=\csname l@#1\endcsname
\fi
#2}}
\providecommand{\BIBdecl}{\relax}
\BIBdecl

\bibitem{10.1093/milmed/usac163}
J.~Larsson and Olsson, ``Physiological demands and characteristics of movement during simulated combat,'' \emph{Military Medicine}, vol. 188, p. 3496, 06 2022.

\bibitem{teyhen2018incidence}
D.~S. Teyhen and Goffar, ``Incidence of musculoskeletal injury in us army unit types,'' \emph{Journal of orthopaedic \& sports therapy}, vol.~48, no.~10, p. 749, 18.

\bibitem{hauret2015epidemiology}
K.~G. Hauret and Bedno, ``Epidemiology of exercise-and sports-related injuries in a population of young, physically active adults,'' \emph{The American journal of sports medicine}, vol.~43, no.~11, p. 2645, 2015.

\bibitem{molloy2020musculoskeletal}
J.~M. Molloy and Pendergrass, ``Musculoskeletal injuries and united states army readiness,'' \emph{Military medicine}, vol. 185, no.~9, p. e1461.

\bibitem{USDOD2023wearables}
\BIBentryALTinterwordspacing
{U.S. Department of Defense}, ``Use of fitness wearables to measure and promote readiness,'' Military Health System, Report, 2023. [Online]. Available: \url{https://health.mil/Reference-Center/Reports/2023/07/24/Use-of-Fitness-Wearables-to-Measure-and-Promote-Readiness}
\BIBentrySTDinterwordspacing

\bibitem{wearableDevicesSailorSleep}
{Sleep Review}, ``Wearable devices monitor sailor sleep during naval exercise,'' \emph{Sleep Review Magazine}, 23, accessed: 2024-12-20.

\bibitem{zhang2022deep}
S.~Zhang, Y.~Li, and Zhang, ``Deep learning in human activity recognition with wearable sensors,'' \emph{Sensors}, vol.~22, no.~4, p. 1476, 2022.

\bibitem{mukherjee2017smartarm}
A.~Mukherjee and Misra, ``Smartarm: A smartphone-based group activity recognition and monitoring scheme for military applications,'' in \emph{IEEE Conference on Advanced Networks Telecommunications}.

\bibitem{khatun2022deep}
M.~A. Khatun and Yousuf, ``Deep cnn-lstm with self-attention model for human activity recognition using wearable sensor,'' \emph{IEEE Journal of Translational Engineering in Health and Medicine}, vol.~10, pp. 1--16, 2022.

\bibitem{ngiam2011multimodal}
J.~Ngiam and Khosla, ``Multimodal deep learning,'' in \emph{28th international conference on machine learning}, 2011, pp. 689--696.

\bibitem{nindl2015human}
B.~C. Nindl and Jaffin, ``Human performance optimization metrics,'' \emph{The Journal of Strength \& Conditioning Research}, vol.~29, p. S221, 2015.

\bibitem{challa2022multibranch}
S.~K. Challa, A.~Kumar, and V.~B. Semwal, ``A multibranch cnn-bilstm model for human activity recognition using wearable sensor data,'' \emph{The Visual Computer}, vol.~38, no.~12, p. 4095, 2022.

\bibitem{zhao2021data}
M.~Zhao, D.~Wang, and J.~Li, ``Data management and visualization of wearable medical devices,'' \emph{Network Modeling Analysis in Health Informatics and Bioinformatics}, vol.~10, no.~1, p.~53, 2021.

\bibitem{mao2023hybrid}
Y.~Mao and Yan, ``A hybrid human activity recognition method using an mlp neural network,'' \emph{Applied Sciences}, vol.~13, no.~18, p. 10529, 2023.

\bibitem{xie2024active}
Z.~Xie, C.~Jiao, and Wu, ``Active factor graph network for group activity recognition,'' \emph{IEEE Transactions on Image Processing}, 2024.

\bibitem{roberts2020detecting}
D.~M. Roberts and Schade, ``Detecting sleep using heart rate and motion data from multisensor consumer-grade wearables,'' \emph{Sleep}, vol.~43, no.~7, p. zsaa045, 2020.

\bibitem{ma2018intelligent}
X.~Ma, Z.~Wang, and Zhou, ``Intelligent healthcare systems assisted by data analytics and mobile computing,'' \emph{Wireless Communications and Mobile Computing}, vol. 2018, no.~1, p. 3928080, 2018.

\bibitem{georgiaTechHeatIllness}
{Signal Processing Society}, ``Georgia tech contributes to reducing heat-related illness in military personnel,'' \emph{IEEE Signal Processing Society Newsletter}, 2020.

\bibitem{gtriWearableSensorSystem}
\BIBentryALTinterwordspacing
{Georgia Tech Research}, ``Gtri helps develop wearable sensor system to prevent heat injuries among soldiers,'' \emph{Georgia Tech Research}, 2020. [Online]. Available: \url{https://research.gatech.edu}
\BIBentrySTDinterwordspacing

\bibitem{luo2021spectral}
X.~Luo, H.~Gao, and Yu, ``Spectral analysis of heart rate variability for trauma outcome prediction,'' \emph{European Journal of Trauma and Emergency Surgery}, vol.~47, pp. 153--160, 2021.

\bibitem{kleinsasser2022detecting}
M.~Kleinsasser and Tharpe, ``Detecting and predicting sleep activity using biometric sensor data,'' in \emph{2022 14th International Conference on COMmunication Systems \& NETworkS}.\hskip 1em plus 0.5em minus 0.4em\relax IEEE, 2022, pp. 19--24.

\bibitem{lovdal2021injury}
S.~L{\"o}vdal \emph{et~al.}, ``Injury prediction in competitive runners,'' \emph{ijspp}, vol. 2020, p. 0518, 2021.

\bibitem{ortiz2024data}
B.~L. Ortiz and Gupta, ``Data preprocessing techniques for ai and machine learning readiness,'' \emph{JMIR mHealth}, vol.~12, no.~1, p. e59587, 2024.

\bibitem{mandadapu2023automated}
\BIBentryALTinterwordspacing
P.~Mandadapu, ``Automated human activity recognition from controlled environment videos,'' Thesis, University of Wisconsin-Milwaukee, Dec 2023. [Online]. Available: \url{dc.uwm.edu/etd/3422}
\BIBentrySTDinterwordspacing

\bibitem{baust1969regulation}
W.~Baust and B.~Bohnert, ``The regulation of heart rate during sleep,'' \emph{Experimental brain research}, vol.~7, pp. 169--180, 1969.

\bibitem{schrack2018using}
J.~A. Schrack and Leroux, ``Using heart rate and accelerometry to define quantity and intensity of physical activity in older adults,'' \emph{Journals of Gerontology}, vol.~73, no.~5, p. 668, 2018.

\bibitem{schlagintweit2023effects}
J.~Schlagintweit, N.~Laharnar, M.~Glos, M.~Zemann, and Demin, ``Effects of sleep fragmentation and partial sleep restriction on heart rate variability,'' \emph{Scientific Reports}, vol.~13, no.~1, 23.

\bibitem{chandola2013sleep}
Chandola, ``Sleep duration and cardiovascular responses to stress in healthy young men,'' \emph{Psychosomatic Medicine}, vol.~75, no.~8, p.~7, 13.

\bibitem{fazli2021hhar}
M.~Fazli and Kowsari, ``Hhar-net: H ierarchical h uman activity r ecognition using neural net works,'' in \emph{Intelligent Human Computer Interaction:IHCI 2020}.\hskip 1em plus 0.5em minus 0.4em\relax Springer, 2021, p.~48.

\bibitem{murtagh2012algorithms}
F.~Murtagh and P.~Contreras, ``Algorithms for hierarchical clustering,'' \emph{Wiley Interdisciplinary: Data Mining and Knowledge Discovery}, vol.~2, no.~1, p.~86, 2012.

\bibitem{yeung2022unified}
M.~Yeung and Sala, ``Unified focal loss: Generalising dice and cross entropy-based losses to handle class imbalanced medical image segmentation,'' \emph{Computerized Medical Imaging and Graphics}, vol.~95, p. 102026, 2022.

\bibitem{ross2017focal}
T.-Y. Ross and G.~Doll{\'a}r, ``Focal loss for dense object detection,'' in \emph{IEEE conference on computer vision and pattern recognition}, 2017, pp. 2980--2988.

\bibitem{stojchevska2023lab}
M.~Stojchevska and D.~Brouwer, ``From lab to real world: Assessing the effectiveness of human activity recognition,'' \emph{Sensors}, 2023.

\bibitem{loshchilov2017fixing}
I.~Loshchilov, F.~Hutter \emph{et~al.}, ``Fixing weight decay regularization in adam,'' vol.~5, 2017.

\bibitem{smith2017cyclical}
L.~N. Smith, ``Cyclical learning rates for training neural networks,'' in \emph{2017 IEEE winter conference on applications of computer vision (WACV)}.\hskip 1em plus 0.5em minus 0.4em\relax IEEE, 2017, pp. 464--472.

\bibitem{murdock2018soldier}
R.~C. Murdock and J.~A. Hagen, ``Soldier safety and performance through wearable devices,'' in \emph{Micro-and Nanotechnology Sensors}.

\bibitem{mongin2020complex}
D.~Mongin and Chabert, ``The complex relationship between effort and heart rate,'' \emph{Physiological measurement}, 2020.

\end{thebibliography}
\end{document}